\pdfoutput=1

\documentclass[11pt]{article}

\usepackage{acl}

\usepackage{times}
\usepackage{latexsym}

\usepackage[T1]{fontenc}

\usepackage[utf8]{inputenc}

\usepackage{microtype}

\usepackage{natbib}
\usepackage{amsmath}
\usepackage{amsfonts}
\usepackage{pifont}
\usepackage{adjustbox}

\title{Tree-Based Representation and Generation of \\ Natural and Mathematical Language}

\author{Alexander Scarlatos \and Andrew Lan \\
        University of Massachusetts Amherst \\
\texttt{\{ajscarlatos,andrewlan\}@cs.umass.edu}}

\begin{document}
\maketitle
\begin{abstract}

Mathematical language in scientific communications and educational scenarios is important yet relatively understudied compared to natural languages. Recent works on mathematical language focus either on representing stand-alone mathematical expressions, especially in their natural tree format, or mathematical reasoning in pre-trained natural language models. Existing works on jointly modeling and generating natural and mathematical languages simply treat mathematical expressions as text, without accounting for the rigid structural properties of mathematical expressions. In this paper, we propose a series of modifications to existing language models to jointly represent and generate text and math: representing mathematical expressions as sequences of node tokens in their operator tree format, using math symbol and tree position embeddings to preserve the semantic and structural properties of mathematical expressions, and using a constrained decoding method to generate mathematically valid expressions. We ground our modifications in GPT-2, resulting in a model MathGPT, and demonstrate that it outperforms baselines on mathematical expression generation tasks. 

\end{abstract}

\section{Introduction}

A part of human communication is performed in rigorous mathematical language rather than more flexible natural language, which often occurs in scenarios such as scientific communication and education. While pre-trained large language models such as BERT \cite{bert} and GPT-3 \citep{brown2020language} have enjoyed many successes in representing and generating natural language, there is a need for models that are effective in representing and generating principled mathematical language as well. While existing work focuses on various aspects of mathematical language representation or generation, combining mathematical language with the aforementioned models for natural language remains a challenging problem. 

Mathematical and natural language are fundamentally different in many ways. While natural language consists of large sets of words and phrases that often have their meaning grounded in context, mathematical language consists of different symbols: a small set of mathematical operators with precise meaning, variables, and numbers that exist in a continuous space. Furthermore, mathematical language follows rules that are much more strict and rigorous than natural language. For example, the multiplication operation acts on exactly two operands, while an integral operates on a single operand but with upper and lower limit arguments. Operands are either variables, numbers, or the result of applying other operations. Given its hierarchical property, mathematical language is naturally represented with operator trees (OPTs), which are directed tree graphs where non-leaf nodes are operators and leaf nodes are variables or numbers \cite{zanibbi2016ntcir,tangentcft}. OPTs are effective at capturing both the semantic and structural properties of mathematical expressions. 

Existing work primarily focuses on two separate approaches to modeling mathematical language: representation and mathematical reasoning. A line of work focuses on learning meaningful representations of mathematical expressions (often formulas), such as \citet{forte,tangent-s}, motivated by the task of retrieving similar expressions, which is especially relevant in information search and retrieval. Although these methods produce dense representations of expressions in their natural tree form, they cannot be directly connected to natural language. Some works employ BERT-like models to jointly represent natural and mathematical language \citep{liang-etal-2022-mwp,mathbertgood,mathbertcrap}. However, these methods are not well suited for generation tasks. 

Another line of work focuses on mathematical reasoning, motivated by the task of mathematical problem solving that is especially relevant in educational applications. These works treat problem solving as a sequence-to-sequence task \cite{saxton2019analysing} and have found success on solving word problems \cite{huang-etal-2018-neural,zou-lu-2019-text2math}. State-of-the-art methods use pre-trained large language models \cite{cobbe2021training,minerva} and can even generate meaningful step-by-step solutions \cite{wei2022chain}. However, these works do not take the principled structure of math into account and treat mathematical expressions as sequences of math LaTeX tokens in the same way as text tokens \cite{taylor2022galactica}. 

\subsection{Contributions}

In this paper, we introduce a series of novel modifications to language models for the joint representation and generation of natural and mathematical languages. We apply these modifications to the publicly available GPT-2 model as a proof-of-concept, although we believe that these modifications apply to many autoregressive language models. Our contributions can be summarized as follows:

\begin{itemize}
\itemsep0em
    \item We develop a set of embeddings that preserve both the semantic and structural properties of mathematical expressions and connect them to natural language token embeddings used in language models. Our embeddings couple the semantic meaning of math tokens with their textual counterparts and explicitly capture the position of nodes in the OPT of an expression.
    \item We develop a parallelizeable constrained decoding procedure that generates mathematically valid expressions via a set of rules.
    \item We apply these modifications to GPT-2 and pre-train it on math Wikipedia articles, resulting in a model we call MathGPT.\footnote{\url{https://github.com/umass-ml4ed/mathGPT}} We demonstrate that it outperforms GPT-2 (and other baselines) on downstream generative tasks, especially on generating math expressions, and analyze how it captures the semantic and structural properties of math expressions using its semantic and position embeddings.
\end{itemize}

\section{Methodology}

\begin{figure}
    \centering
    \includegraphics[width=3in]{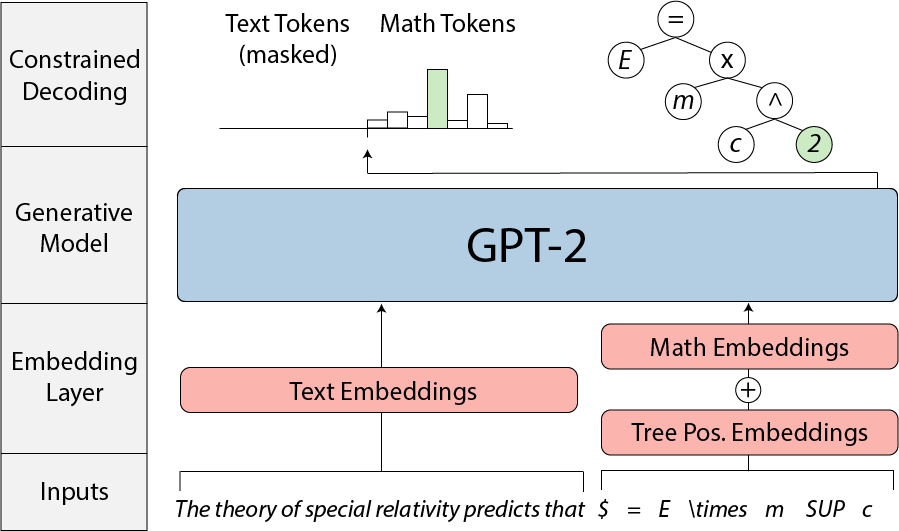}
    \vspace{-.4cm}
    \caption{We represent text and math regions differently, with tree position embeddings added to the math token embeddings. The predicted token probability distribution is shown for the next token, \textit{2}; text tokens are masked out by decoding constraints.}
    \label{fig:arch}
\end{figure}

We now detail our proposed modifications grounded in our model MathGPT, visualized in Figure \ref{fig:arch}. First, we detail how natural and mathematical language are represented jointly by the model. Second, we detail how we provide the model with token-level tree position information, followed by how we represent math token embeddings via a learnable transformation on text token embeddings. Third, we detail our rules for constrained decoding and tree structure inference at test time.

\subsection{Sequence Representation}

We consider sequences that contain separable regions of text and math, i.e., $S = ( T_1, F^{s}, M_1, F^{e}, T_2, F^{s}, M_2, F^{e}, \ldots )$, where $T_n = ( t_1, \ldots, t_N )$ is a sequence of text tokens, $M_n = ( m_1, \ldots, m_N )$ is a sequence of math tokens, $F^{s}$ indicates the start of a mathematical expression, and $F^{e}$ indicates the end of an expression. To leverage the structural information of the expressions, we convert each $M_n$ into its corresponding OPT, $M^{tree}_n$. In $M^{tree}_n$, each token $m_i$ is assigned a node in the tree, and each $m_i \in \{ \mathcal{O}, \mathcal{V}, \mathcal{N}, E \}$, where $\mathcal{O}$ is the set of all operators, $\mathcal{V}$ is the set of all variables, $\mathcal{N}$ is the set of all numbers, and $E$ is a special \textit{end} node. In the tree, operators become parent nodes and their children are either variables, numbers, or other operators. After this initial conversion, we make several modifications to the tree to assist the model with mathematical understanding. First, we add an $E$ node as the last child of every operator node to indicate the end of its list of children. Second, we convert each number into a sub-tree where the head is a special operator $O^N$ and its children are the digits of the number, including the decimal point. Third, since we use a fixed-size vocabulary for math tokens, any out-of-vocabulary token $m_i$ is converted into a sub-tree where the head is a special operator $O^U$ and its children are the text tokens of $m_i$. Then, we convert $M^{tree}_n$ back to a linear sequence, $M^{lin}_n$, by traversing and adding nodes in depth-first order, resulting in the sequence $S^\prime = ( T_1, F^{s}, M^{lin}_1, F^{e}, T_2, F^{s}, M^{lin}_2, F^{e}, \ldots )$. See Supplementary Material \ref{pipeline-example} for an illustration of this process along with other data processing steps.

To convert each token $s_i \in S^\prime$ to its embedding $\mathbf{s}_i$ as input to a language model, GPT-2 follows
\begin{align*}
    \mathbf{s}_i = \operatorname{emb}_{tok}(s_i) + \operatorname{emb}_{sp}(i),
\end{align*}
where $\operatorname{emb}_{tok}(s_i)$ is the token embedding for $s_i$, and $\operatorname{emb}_{sp}(i)$ is the position embedding at index $i$. Our key innovation in MathGPT is a set of modified position embeddings that explicitly provide the model with OPT structure information: 
\begin{align*}
    \mathbf{s}_i = & \operatorname{emb}_{tok}(s_i) + \operatorname{emb}_{sp}(i)\\
    + & \operatorname{emb}_{tp}(\mathbf{p}_i) + \operatorname{emb}_{type}(s_i),
\end{align*} 
where $\operatorname{emb}_{type}(s_i)$ is the symbol \textit{type} embedding (text, operator, variable, etc.) of $s_i$ and $\operatorname{emb}_{tp}(\mathbf{p}_i)$ is the \textit{tree position} embedding of $s_i$. Our approach explicitly captures the tree position and role of each math symbol in the context of the entire mathematical expression to preserve both semantic and structural properties. We also use different \textit{semantic} embeddings $\operatorname{emb}_{tok}(s_i)$ for text and math tokens, which we detail below.  

\subsection{Tree Position Encoding}

For tree position embeddings, we first define $\mathbf{p}_i$, a unique position vector for node $m_i$, and then use a function $\operatorname{emb}_{tp}$ to transform $\mathbf{p}_i$ to a vector with the same dimensionality as the token embeddings. We encode tree positions similar to the approach in \citep{forte}: $\mathbf{p}_i$ is a vector where the entry at each index, $p_i^j$, represents $m_j$'s index in its list of siblings. By following the indices in $\mathbf{p}_i$ from the tree root, node $m_i$ will eventually be reached, and its index is the last entry in the vector. We then convert $\mathbf{p}_i$ to a vectorized version of the binary representation of each of its entries and finally project the resulting vector using a learnable transformation. The whole process is defined as
\begin{align*}
    & bin(p_i^j) = \operatorname{concat}(onehot(b_j^1),\ldots,onehot(b_j^K)) \\
    & bin(\mathbf{p}_i) = \operatorname{concat}(bin(p_i^1),\ldots,bin(p_i^{|\mathbf{p}_i|}))  \\
    & \operatorname{emb}_{tp}(\mathbf{p}_i) = \mathbf{W}bin(\mathbf{p}_i),
\end{align*}
where $b_j^k$ is the $k$th digit of $p_i^j$'s binary representation, $onehot$ returns a one-hot 2-vector, and $\mathbf{W}$ is a learnable projection matrix.

\subsection{Math Token Embeddings}

We construct our semantic embeddings for the math symbols by linking them with the corresponding text tokens in GPT-2's pre-trained vocabulary. Specifically, let the text representation of a math symbol $s_i$ be $t_i$. We tokenize $t_i$ with the GPT-2 tokenizer to produce a corresponding sequence of text tokens, $(t_i^1, \ldots, t_i^K)$. The embedding of $s_i$ is then given by
\begin{align*}
    & \mathbf{t}_i = \textstyle\sum^K_{k=1}{\operatorname{emb}_{tok}(t_i^k)} / K \\
    & \operatorname{emb}_{tok}(s_i) = \mathbf{t}_i + \phi_p(\mathbf{t}_i),
\end{align*}
where $\phi_p$ is a fully-connected neural network with a single hidden layer, and we initialize the weights to be small such that $\operatorname{emb}_{tok}(s_i)$ is initially very close to $\mathbf{t}_i$. With this formulation, we leverage the pre-trained information in GPT-2 while updating text token representations during training through MathGPT's tree-structured representations for mathematical expressions. For math symbols that have no corresponding text representations, such as $F^s$, $F^e$, $O^N$, and $O^U$, we learn their semantic embeddings from scratch. 

\subsection{Sequence Generation}

In addition to modifying GPT-2's input format, we also make several changes to the output process to enable MathGPT to generate mathematically meaningful expressions. We create a new linear predictor head for math tokens, including special control tokens ($F^s$, $F^e$, etc.), $\phi_{math}$. We concatenate the output of this projection to those of the pre-trained text prediction head, $\phi_{text}$, to get a full token probability vector, $\mathbf{a}_i$, at each time step.

To ensure that MathGPT generates mathematically valid expressions, we employ constrained decoding by applying a mask to $\mathbf{a}_i$ that prohibits certain tokens from being generated after $s_i$. We apply the following constraints: First, text tokens must be followed by text tokens or $F^s$. Second, $F^s$ must be followed by operator, variable, or number tokens. Third, $F^e$ must be followed by text tokens. Fourth, operator, variable, number, and $E$ tokens must be followed by other operator, variable, number, or $E$ tokens. The exception is when a tree has been fully generated, in which case they must be followed by $F^e$. Fifth, trees have limited depth and width, so we prevent operator nodes from being generated at the maximum depth level and cap the maximum number of children for each node. Finally, $O^U$ must be followed by text tokens, which can be followed by other text tokens or $E$, and $O^N$ must similarly be followed by number tokens.

During training, we minimize the cross-entropy loss using the masked version of token probabilities $\textbf{a}_i$ to update the GPT-2 parameters along with the MathGPT-specific parameters, including $\phi_{math}$, $\phi_p$, $\mathbf{W}$, and embeddings of the special tokens.

During testing (generation), we infer the tree position of the next node directly from the position of the previous node from depth-first ordering, according to the following rules:
If $s_i \in \mathcal{O}$, then $s_{i+1}$ will be its first child. Thus $\mathbf{p}_{i+1}$ will be a copy of $\mathbf{p}_i$ with a $0$ added to the end. If $s_i \in \{\mathcal{V}, \mathcal{N}\}$, then $s_{i+1}$ will be its next sibling. Thus $\mathbf{p}_{i+1}$ will be a copy of $\mathbf{p}_i$ where the last value is incremented by 1. If $s_i = E$, then $s_{i+1}$ will be its parent's sibling. Thus $\mathbf{p}_{i+1}$ will be a copy of $\mathbf{p}_i$ without the last value and the preceding value incremented by 1.

\section{Experimental Setup}

We now detail a series of experiments to validate the effectiveness of MathGPT. We perform pre-training on a large corpus of math-focused Wikipedia articles and then use the model on various generative downstream tasks involving both natural and mathematical languages. 

\subsection{Data Pre-Processing}

In the pre-training and downstream task datasets, the mathematical expressions are initially represented as plain text or HTML, occasionally wrapped in text-based $F^s$ and $F^e$ tokens, and pre-converted to MathML in the pre-training dataset. To convert them to their OPT representations, we introduce the following data pre-processing pipeline. 
First, we convert all HTML math-specific symbols, including variables, numbers, and operators, to their LaTeX equivalents, and remove remaining HTML tags. Second, we find all expressions in each text sequence and wrap them with $F^s$ and $F^e$ tokens. Third, we process each sequence with LaTeXML \footnote{\url{https://math.nist.gov/~BMiller/LaTeXML/}}, which converts each expression to a tree-like MathML representation. Finally, we process each MathML expression with code from Tangent-CFT \citep{tangentcft} to obtain its standard OPT representation. 

We note that LaTeXML introduces several undesirable distortions on mathematical expressions. For example, it is often unable to differentiate between function calls and multiplications with parentheses, multi-character names and multiplications between single character variables, numbers containing commas and comma-delimited lists of numbers, etc. However, we found that in the majority of cases it is accurate enough. 

\subsection{Pre-Training}

We use a pre-trained GPT-2 model to initialize the shared parameters in MathGPT, which enables us to leverage GPT-2's existing representations and capabilities. We then pre-train MathGPT on a large corpus of math-centered Wikipedia articles from the \textit{2016 NTCIR-12 MathIR Task} \citep{ntcir-12},
which enables the model to learn the parameters that are unique from GPT-2. We reserve 5\% of the articles for validation and pre-train for 50 epochs, which we found to be sufficient for the model to perform reasonably well on downstream tasks.
Additional hyperparameters and model details are listed in Supplementary Material \ref{hyperparams}.

\subsection{Downstream Tasks}

We evaluate MathGPT on the following downstream generative tasks, which together capture a wide range of mathematical reasoning capabilities.
Additional details on the datasets can be found in Supplementary Material \ref{appendix-datasets}.

\paragraph{Headline Generation} We evaluate on the task of math headline generation using the EXEQ-300k dataset \citep{mathsum}, which contains pairs of user-authored questions and headlines from Mathematics Stack Exchange. The content in this dataset is generally on college-level math and science topics, containing complex formulas with a large variety of symbols. This task measures the model's ability to extract key information from the question and generate a short summary. Due to reasons we detail below in Section~\ref{sec:metrics}, we additionally consider two sub-tasks: next mathematical expression prediction and next text region prediction. For next mathematical expression prediction, we consider each expression in the headline to be a generation target, while we use both the question and the portion of the headline up to that expression as input. Similarly, for next text region prediction, we consider each text region that follows a mathematical expression to be a generation target, while we use both the question and the portion of the headline up to that text region as input. 

\paragraph{Equation Extraction} We evaluate on the task of generating equations in math word problems on a version of the Math23K \citep{math23k} dataset converted to English using Google Translate. While the translations are not perfect, we do see that they largely retain the necessary mathematical information. The dataset contains pairs of middle school-level math word problems and single-variable equations that represent an execution plan to solve the problem. This task measures the model's ability to infer mathematical operations and expression structure from unstructured text.

\paragraph{Student Action Prediction} We evaluate on the task of predicting how students act based on feedback while solving math problems in a step-by-step setting. We use a dataset \footnote{\url{https://pslcdatashop.web.cmu.edu/DatasetInfo?datasetId=660}} from the Cognitive Tutor system. At each step, the student chooses an action (add, subtract, multiply, etc.) and enters a corresponding input (a mathematical expression) to perform on the problem's equation. If the action is incorrect, the system will provide feedback to the student and let them retry. Our task is to predict exactly what actions students make after receiving feedback after incorrect steps, using the equation being solved, a series of steps the student made, sequential updates to the equation, and a feedback message as input to generate the following student action, input, and outcome as output. This task measures the model's ability to predict which action a student will take based on their previous actions, which involves knowing what the appropriate next step is for solving an equation.

\subsection{Evaluation Metrics}
\label{sec:metrics}

Since we evaluate MathGPT on a variety of tasks with different objectives, we similarly measure performance using a wide set of task-specific metrics. For headline generation, where we measure the quality of both generated math and natural language, we use text similarity metrics including \textbf{BLEU-4} \citep{bleu}, \textbf{ROUGE-L} \citep{rouge}, and \textbf{METEOR} \citep{meteor}. However, since MathGPT outputs mathematical expressions as OPTs while the baselines output them as a sequence of LaTeX tokens, we convert MathGPT's expression output back to LaTeX using a custom tree parser before computing these metrics. We compare the generated output for MathGPT and baselines to a modified version of the ground truth, where the expressions are converted to OPTs via LaTeXML and then converted back to text via the parser. This conversion is necessary since LaTeXML can change the semantics of an expression.

However, these metrics are insufficient since they do not consider the structural integrity of math expressions; for MathGPT, the expressions are generated as trees yet evaluated as text token sequences. To the best of our knowledge, there is no automated metric that can effectively evaluate natural and mathematical languages jointly. Additionally, while human evaluation can be valuable, designing such an experiment is challenging since we need to account for individual text and math properties as well as cohesion between them. We leave both of these aspects for future work. In the current paper, we circumvent this roadblock by including two new evaluation tasks that evaluate text and math \emph{separately}. On math expressions, we use tree edit distance (\textbf{TED}) to evaluate their structural integrity.

We use pre-defined train/validation/test splits on the headline generation dataset, and report mean and standard deviation for each metric on the test set over 5 random initializations. For other downstream datasets where pre-defined splits are not available, we perform a 5-fold cross-validation, where the train/test sets are rotated and 10\% of the remaining train set is reserved for validation. We similarly report the mean and and standard deviation of each metric on the test set over the 5 folds. In all cases, we perform early stopping on per-token loss on the validation set. In all result tables, we place a * next to a metric value for MathGPT if it outperforms baselines with statistical significance, i.e., $p < 0.05$ from the Student's t-test for cross-validation and Welch's t-test otherwise.

\subsection{Baselines}

Since the key innovation in MathGPT is a structural modification on top of the original GPT-2 model, our goal is to show that MathGPT outperforms GPT-2 in terms of representing and generating mathematical content. Therefore, we use i) standard GPT-2 and ii) GPT-2 pre-trained on the math-centric Wikipedia articles as our baselines. Moreover, for some of the downstream tasks, we also report state-of-the-art results as an additional baseline. For a fair comparison with MathGPT on text-based metrics, for the headline generation task, we train and evaluate GPT-2 on a version of the dataset where the mathematical expressions are converted to OPTs via the processing pipeline and then back to LaTeX using the tree parser. For all other tasks, we train GPT-2 on the original dataset.

\section{Experimental Results}

We now detail quantitative experimental results to validate the effectiveness of MathGPT in jointly modeling natural and mathematical languages. 

\subsection{Headline Generation}

\begin{table}
    \centering
    \scalebox{.80}{
    \begin{tabular}{|l|c|c|c|}
        \hline
        \textbf{Model} & \textbf{BLEU-4} & \textbf{ROUGE-L} & \textbf{METEOR} \\
        \hline
        MathBERT & $49.4$ & $57.7$ & $34.7$ \\
        MathSum & $52.0$ & $54.8$ & $37.5$ \\
        \hline
        GPT-2 & $55.3 \pm 1.1$ & $62.1 \pm 0.0$ & $43.7 \pm 0.3$ \\
        GPT-2 Wiki & $56.1 \pm 0.6$ & $62.2 \pm 0.1$ & $43.7 \pm 0.3$ \\
        \hline
        MathGPT & $56.5 \pm 0.5$ & $62.2 \pm 0.1$ & $43.8 \pm 0.3$ \\
        \hline
    \end{tabular}}
    \vspace{-.2cm}
    \caption{Results on headline generation.}
    \label{tab:headline-results}
\end{table}

Table \ref{tab:headline-results} shows results on overall headline generation on the EXEQ-300k dataset; see Supplementary Material \ref{appendix-ofeq} for results on the smaller OFEQ-10k dataset. Table \ref{tab:headline-subtask-results} shows results on the next math expression and text region prediction sub-tasks. We emphasize that when evaluated on text and math regions separately, MathGPT significantly outperforms GPT-2, especially on TED, which captures the structural integrity of math expressions, although part of the reason for GPT-2's high TED numbers can be attributed to occasional parsing errors in LaTeXML. Interestingly, MathGPT also significantly outperforms both GPT-2 models in next text region prediction on all metrics, which suggests that trees are highly effective at conveying the underlying meaning of math expressions, which are reflected in the text regions of the headlines. 

However, when we evaluate text and math regions jointly on existing text-based metrics, MathGPT's advantage over GPT-2 is minimal and not statistically significant. This result can be attributed to the lack of existing metrics that consider the structural properties of math expressions while combining them with text. Both MathGPT and GPT-2 significantly outperform prior state-of-the-art: MathSum, a sequence-to-sequence method and MathBERT \citep{mathbertgood}, a BERT-based method that leverages tree information and adapted to this task. These results show that the GPT family of language models are well-suited to generation tasks on math content even without task-specific architectures such as the copying mechanism.

\begin{table*}
    \centering
    \scalebox{.80}{
    \begin{tabular}{|l|c|c|c|c|c|c|c|}
        \hline
         & \multicolumn{4}{|c|}{\textbf{Next Mathematical Expression}} & \multicolumn{3}{|c|}{\textbf{Next Text Region}} \\
        \hline
        \textbf{Model} & \textbf{BLEU-4} & \textbf{ROUGE-L} & \textbf{METEOR} & \textbf{TED} & \textbf{BLEU-4} & \textbf{ROUGE-L} & \textbf{METEOR} \\
        \hline
        GPT-2 & $77.4 \pm 0.2$ & $83.1 \pm 0.0$ & $56.1 \pm 0.1$ & $4.125 \pm 0.035$ & $42.5 \pm 0.2$ & $58.1 \pm 0.3$ & $38.3 \pm 0.2$ \\
        GPT-2 Wiki & $77.4 \pm 0.2$ & $83.5 \pm 0.1$ & $56.2 \pm 0.0$ & $4.079 \pm 0.045$ & $43.8 \pm 0.5$ & $54.5 \pm 0.2$ & $40.6 \pm 0.0$ \\
        \hline
        MathGPT & $77.6 \pm 0.2$ & *$83.7 \pm 0.1$ & *$56.4 \pm 0.1$ & *$2.656 \pm 0.023$ & *$46.2 \pm 0.1$ & *$63.3 \pm 0.1$ & *$42.2 \pm 0.1$ \\
        \hline
    \end{tabular}}
    \caption{Results on next mathematical expression (left) and next text region (right) prediction.}
    \label{tab:headline-subtask-results}
\end{table*}

\subsection{Underlying Equation Extraction}

\begin{table}
    \centering
    \scalebox{.80}{
    \begin{tabular}{|l|c|c|c|}
        \hline
        \textbf{Model} & \textbf{Tree Match} & \textbf{Solve Rate} & \textbf{TED} \\
        \hline
        GPT-2 & $47.8 \pm 1.0$ & $54.6 \pm 1.1$ & $2.669 \pm 0.099$ \\
        GPT-2 Wiki & $47.2 \pm 0.7$ & $54.0 \pm 0.8$ & $2.595 \pm 0.024$ \\
        \hline
        MathGPT & *$52.4 \pm 1.1$ & *$60.3 \pm 1.3$ & $2.449 \pm 0.098$ \\
        \hline
    \end{tabular}}
    \caption{Results on equation extraction.}
    \label{tab:eq-results}
\end{table}

Table \ref{tab:eq-results} shows results on the equation extraction task. We also include two task-specific metrics: the percentage of cases where the generated equation and the true equation have the same exact same OPT (\textbf{Tree Match}), and the percentage of cases where both evaluate to the same numerical value (\textbf{Solve Rate}). We see that MathGPT outperforms both GPT-2 models significantly on all metrics, which implies that MathGPT is effective at both extracting mathematical information from textual problem statements and generating equations as the solution. 
We observe that MathGPT's advantage over GPT-2 on TED is less than that on the other metrics, due to GPT-2 sometimes generating trees that are similar to the ground truth but containing a few key errors such as using an incorrect operator. We also observe that Solve Rate is higher than Tree Match for all models, since models often generate equations that evaluate to the correct numerical value but have slightly different trees.

\subsection{Student Action Prediction}

\begin{table}
    \centering
    \scalebox{.89}{
    \begin{tabular}{|l|c|c|}
        \hline
        \textbf{Model} & \textbf{Accuracy} \\
        \hline
        GPT-2 & $40.0 \pm 0.8$ \\
        GPT-2 Wiki & $40.3 \pm 1.2$ \\
        \hline
        MathGPT & *$41.8 \pm 1.0$ \\
        \hline
    \end{tabular}}
    \caption{Results on student action prediction.}
    \label{tab:ct-results}
\end{table}

Table \ref{tab:ct-results} shows results on the student action prediction task where we only report the prediction \textbf{Accuracy} on each (outcome, action, input) triple. We observe that MathGPT outperforms both GPT-2 models. More specifically, when students are correct, MathGPT predicts the action and input correctly 63.5\% of the time, whereas GPT-2 and GPT-2 with math Wikipedia pre-training are correct 61.2\% and 61.5\% of the time, respectively. However, when students are incorrect, these numbers significantly decrease to 6.6\%, 6.4\%, and 6.8\%. This observation implies that MathGPT outperforms GPT-2 on mathematical reasoning but not on predicting student errors, which is expected since these models do not account for variation in student knowledge.

\subsection{Ablation Study}

\begin{table}
    \centering
    \scalebox{.70}{
    \begin{tabular}{|c|c|c|c|c|c|c|c|}
        \hline
        \textbf{TPE} & \textbf{TE} & \textbf{SE} & \textbf{NT} & \textbf{Tree Match} & \textbf{TED} & \textbf{Solve Rate} \\
        \hline
        \hline
        \multicolumn{7}{|c|}{Pre-trained for 50 epochs}\\
        \hline
         & & & & $52.4 \pm 1.1$ & $2.449 \pm 0.098$ & $60.3 \pm 1.3$ \\
        \hline
        \hline
        \multicolumn{7}{|c|}{Pre-trained for 25 epochs}\\
        \hline
         & & & & $53.1 \pm 0.9$ & $2.367 \pm 0.060$ & $61.2 \pm 1.0$ \\
        \ding{53} & & & & $52.6 \pm 0.5$ & $2.371 \pm 0.038$ & $60.3 \pm 0.8$ \\
         & \ding{53} & & & $53.0 \pm 0.4$ & $2.362 \pm 0.021$ & $61.0 \pm 0.7$ \\
         & & \ding{53} & & $51.0 \pm 0.4$ & $2.464 \pm 0.037$ & $58.6 \pm 0.6$ \\
         & & & \ding{53} & $49.4 \pm 0.8$ & $1.734 \pm 0.026$ & $57.1 \pm 1.0$ \\
        \hline
    \end{tabular}}
    \caption{Results of ablation on equation extraction.}
    \label{tab:ablation}
\end{table}

We examine the impact of various components of MathGPT on its downstream performance via an ablation study. Specifically, we create several versions of the model: with no tree position embeddings (\textbf{TPE}), with no math symbol type embeddings (\textbf{TE}), learning unique math token embeddings instead of linking with text token embeddings (\textbf{SE}), and treating most frequent numbers as their own token instead of as subtrees (\textbf{NT}).
We note that it is difficult to ablate on constrained decoding since it is central to MathGPT; without these constraints, we may generate unparseable sequences that cannot be interpreted as trees, making some evaluation metrics invalid (e.g., TED).
We pre-train all models in the ablation study for 25 epochs and evaluate on the equation extraction task.

Table \ref{tab:ablation} shows results for the ablation study. We see that all components, except for type embedding, are critical to downstream task performance. Tree position embeddings have a higher impact on accuracy than TED, likely due to these embeddings helping place nodes in correct positions. Removing numeric sub-trees hurts accuracy, likely since it makes it harder for the model to differentiate between multi-token numbers. It also reduces TED, as expected, since number mismatches have a lower overall penalty. Finally, MathGPT pre-trained for 25 epochs outperforms 50 epochs, which implies the model overfits on the Wikipedia data. This observation suggests a more diverse pre-training dataset would help, which we leave for future work.

\section{Qualitative Analysis}

We now qualitatively investigate how MathGPT represents the semantic and structural aspects of mathematical language differently than GPT-2 fine-tuned on the same math Wikipedia articles.

\subsection{Math Token Embeddings}

\begin{figure}
    \centering
    \includegraphics[width=2.9in]{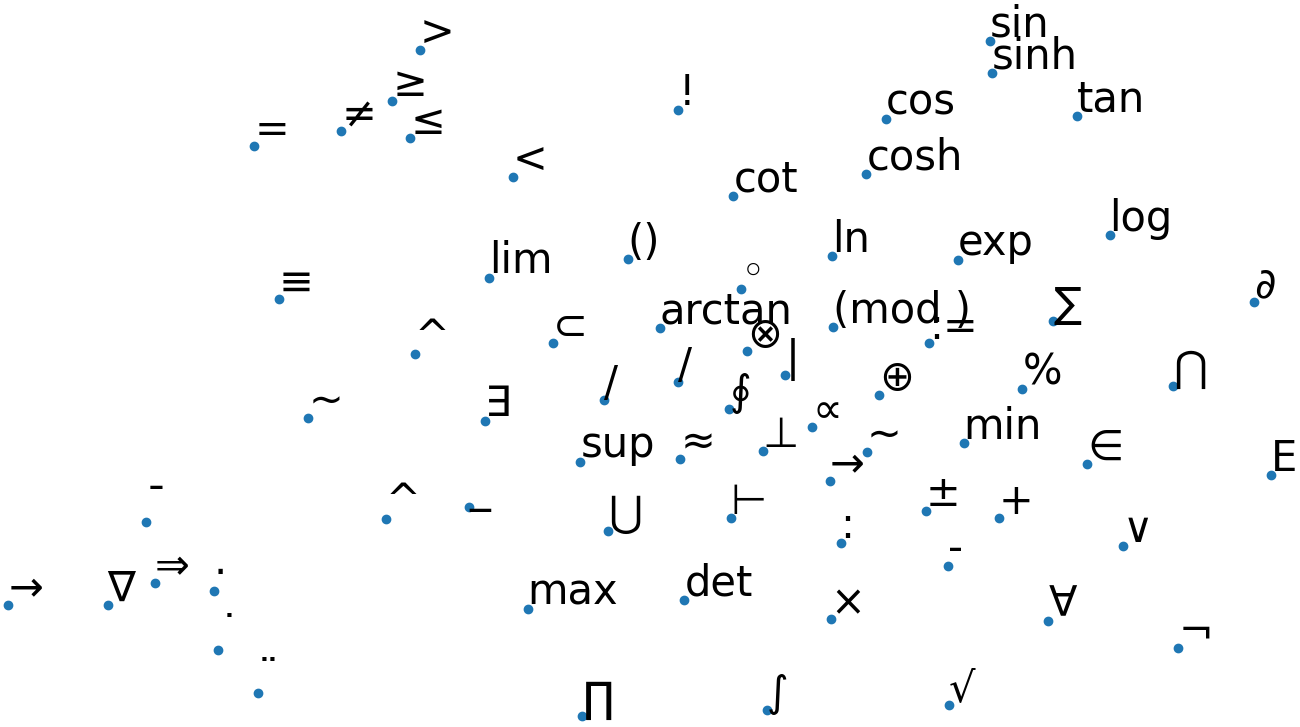}
    \caption{MathGPT operator token embeddings.}
    \label{fig:tokens-mathgpt}
\end{figure}

\begin{figure}
    \centering
    \includegraphics[width=2.9in]{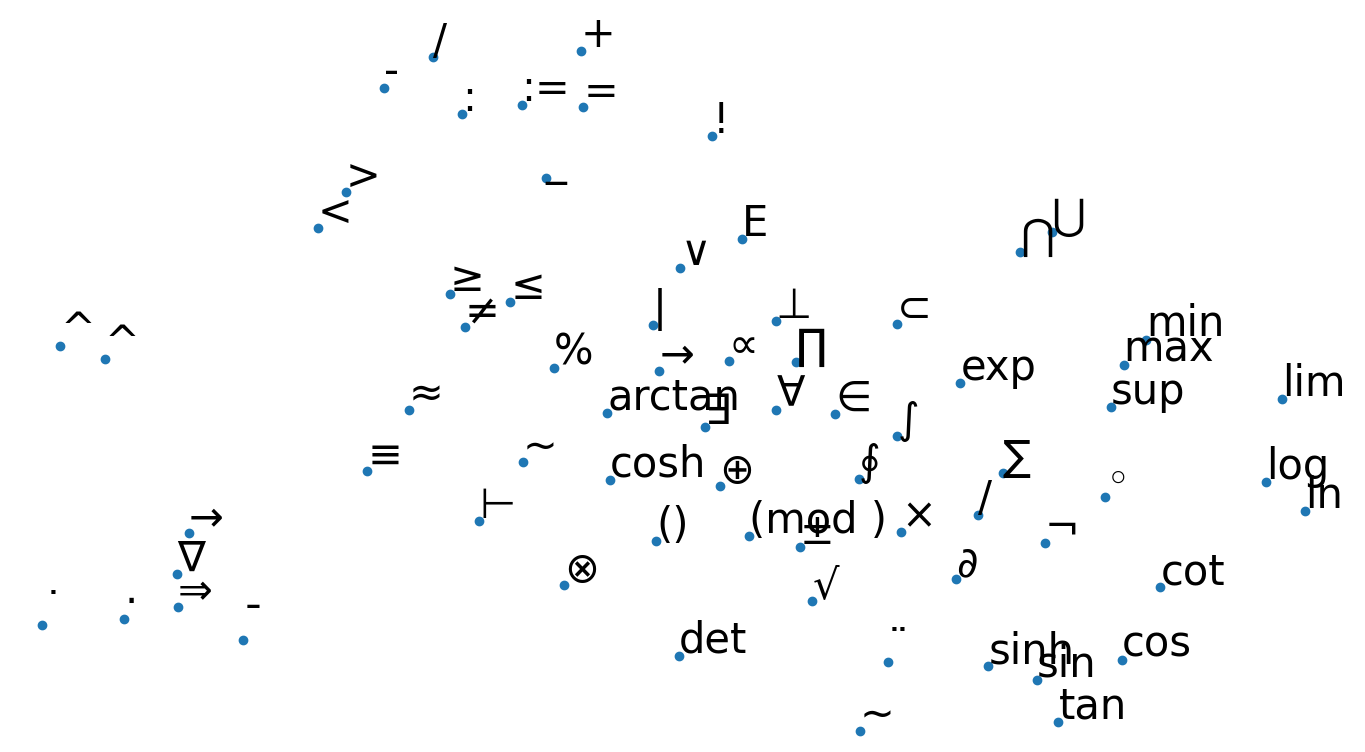}
    \caption{GPT-2 operator token embeddings.}
    \label{fig:tokens-gpt2}
\end{figure}

Figures \ref{fig:tokens-mathgpt} and \ref{fig:tokens-gpt2} show the semantic embeddings of the top 100 most frequent mathematical \emph{operator} tokens for MathGPT and GPT-2, respectively, visualized in 2D using t-SNE \citep{t-sne}. For MathGPT we show $\operatorname{emb}_{tok}(s_i) + \operatorname{emb}_{type}(s_i)$, and for GPT-2 we show the average token embeddings of an operator's LaTeX symbol. We see a few key differences. First, MathGPT seems to group symbols together based on mathematical semantic similarity, whereas GPT-2 seems to group symbols together that may appear in similar contexts. For example, MathGPT groups $=$ and inequalities together and keeps $+$, $-$, and $\pm$ in a separate group. GPT-2 groups $=$ with algebraic operators and other symbols and keeps inequalities in a separate group. Second, MathGPT separates several pairs of symbols that are grouped together by GPT-2 such as ($\min$, $\max$) and ($\bigcup$, $\bigcap$). This observation shows that MathGPT places high importance on an operator's effect on other symbols in addition to its category. We note that different initializations of t-SNE result in different visualizations; see Supplementary Material \ref{appendix-tokens} for details.

\subsection{Tree Position Representations}

\begin{figure}
    \centering
    \includegraphics[width=.4\linewidth]{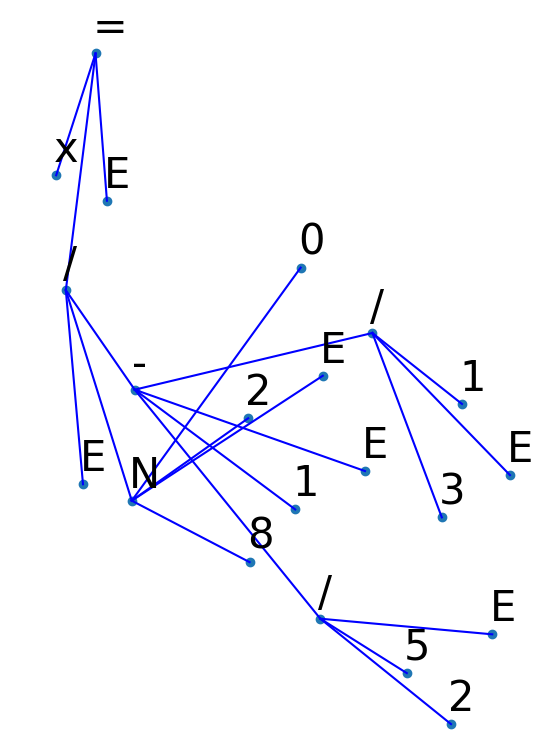}
    \hspace{.1\linewidth}
    \includegraphics[width=.35\linewidth]{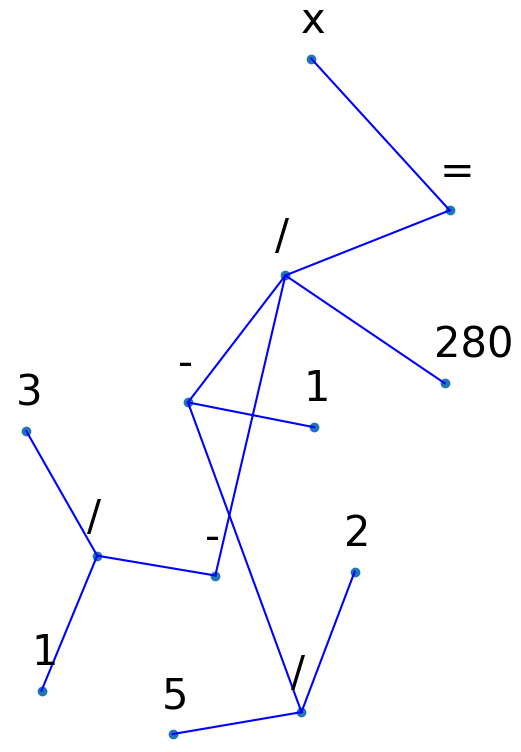}
    \caption{Position embeddings for MathGPT (left) and GPT-2 (right).}
    \label{fig:tpe}
\end{figure}

Figure \ref{fig:tpe} shows the learned (tree) positional embeddings for MathGPT, i.e., $\operatorname{emb}_{sp}(i) + \operatorname{emb}_{tp}(\mathbf{p}_i)$, and for GPT-2, i.e., $\operatorname{emb}_{sp}(i)$, visualized in 2D using t-SNE, for the mathematical expression $x=280/(1-(2/5)-(1/3))$. We see that the MathGPT embeddings clearly show a tree structure, where nodes that are deeper in the tree are further from nodes high in the tree and sibling and cousin nodes are close together. For GPT-2, while nodes that appear later in the expression are far from those that appear early, the mathematical structure of the expression is not clearly reflected. While these results are expected, they show that MathGPT's position embeddings explicitly capture the structural properties of mathematical expressions.

\section{Related Work}
\label{sec:rw}

\noindent{\bf Representations of mathematical language} Existing work on studying the representations of mathematical language is mainly motivated by information \emph{retrieval}, i.e., retrieving a semantically and/or structurally relevant mathematical expression (often formula) given a query \cite{zanibbi2016ntcir,tangent-s}. Both representations based on expert-crafted rules \cite{approach0,approach0-new} and those learned from large-scale scientific formula data \cite{tangentcft,forte} have been shown to be highly effective at this task. However, most of these works do not consider the important textual context around expressions. Several recent works jointly model text and math: TopicEq \cite{topiceq} learns topic keywords associated with expressions, MathSum \cite{mathsum} generates headlines for mathematical discussion forum posts, and one of the MathBERT models \cite{mathbertcrap} learns how to grade students' open-ended math responses. However, none of these works leverage the tree structure of math expressions. Another MathBERT model \cite{mathbertgood} and a recent work \cite{fuse} jointly represent textual tokens and expressions in their tree format. However, neither is naturally suited for \emph{generation} tasks. 

\noindent{\bf Mathematical reasoning in language models} Existing work on studying the mathematical reasoning ability of language models \cite{neural-math-1,neural-math-2} is mainly motivated by the math \emph{problem solving} task. Despite evidence that pre-trained neural language models have limited mathematical reasoning ability \cite{saxton2019analysing,jin2021numgpt}, they are able to solve simple math word problems accurately using techniques such as verifiers \cite{cobbe2021training} and chain-of-thought prompting \cite{wei2022chain}. However, these models do not take the tree structure of mathematical expressions into account and simply represent them as LaTeX token sequences.  

\section{Conclusions and Future Work}
In this paper, we proposed a series of modifications to common language models to represent and generate text and math jointly. We applied these modifications to GPT-2, resulting in the MathGPT model, which excels at capturing the semantic and structural properties of mathematical expressions in generative tasks. There are many avenues of future work, including i) expand the pre-training dataset for MathGPT to cover a wider range of mathematical topics and complexity levels, ii) develop representations of expressions that are invariant under structural transformations that do not change their semantic meaning, iii) conduct human evaluation to further validate the quality of the generated mathematical expressions in multiple aspects and iv) develop similarity metrics for mathematical language (such as CodeBLEU \cite{codebleu} for code) and validate these metrics with evaluations from human experts.

\section*{Limitations}

There are several limitations to MathGPT, both practical and fundamental. First, the model depends on an external method for converting mathematical expressions to OPTs, currently being LaTeXML. The conversion method is imperfect, which limits MathGPT's capabilities as it will be presented with many distorted expressions during training and at test time. Furthermore, the conversion process is slow and requires dataset-specific engineering to accommodate, making it difficult to deploy the model across many datasets. Second, because MathGPT outputs trees rather than sequences, it is fundamentally difficult to evaluate and utilize in text-based settings without a highly accurate tree-to-text converter. The tree-to-text converter is yet another imperfect process in the pipeline, although it could be improved to a reasonable degree with significant engineering effort. Third, because MathGPT has additional components and requires more information per token than GPT-2, it has higher space and time requirements that make training more expensive. Finally, because MathGPT is pre-trained on highly formal and structured mathematical content, it may struggle to generalize to student-generated mathematical language, which is often error-prone and may exhibit very different patterns.

\section*{Ethics Statement}

All large language models are prone to reflecting biases seen in their training data. Since MathGPT would likely find its greatest use in an educational setting, extensive care would have to be taken to identify and mitigate bias against students across demographics and backgrounds if deployed in these settings. It is also possible that different patterns exist in mathematical language written by students across demographic groups, such as the choice of variable names or structural choices that reflect different educational backgrounds. Before being deployed in an educational setting, studies should be performed to ensure that the model would not ``prefer'' any patterns that tend to be exhibited by certain groups of students. It would also be necessary to examine the impact of bias mitigation strategies on removing such preferences and on the effectiveness of the model overall. We did not perform any such studies as part of this work since we did not use any student-generated datasets that contained demographic information, though we welcome such studies and consider it an important part of the future of this line of work.

\section*{Acknowledgement}
The authors are partially supported by the NSF (IIS-2118706, IIS-2202506), the IES (R305N210064), and Schmidt Futures/Walton Family Foundation. We also thank Nigel Fernandez and Zichao (Jack) Wang for helpful discussions around this work. 

\bibliography{anthology,references}

\appendix

\section{Hyperparameters and Implementation Details}
\label{hyperparams}

MathGPT is implemented in PyTorch, using the pre-trained small OpenAI GPT-2 model \citep{radford2019language} from the \textit{HuggingFace Transformers} library \citep{wolf-etal-2020-transformers} as a base. For both pre-training and fine-tuning, we use sequence lengths of 1,024, and we limit OPTs to a depth of 32 and a per-node child count of 64. While multi-digit numbers are converted to sub-trees, we use individual nodes to represent single-digit numbers. We use the AdamW optimizer with a learning rate of 1e-5, a weight decay of 1e-2, and a batch size of 4, accumulating gradients every 4 batches. At test time, we generate sequences using beam search with a width of 3. These hyperparameters were chosen based on exploratory evaluations and are mostly the defaults, and no substantive hyperparameter search was performed. We use the same hyperparameters for training MathGPT and GPT-2. All models were trained on \textit{NVIDIA Quadro RTX 8000} or \textit{NVIDIA Tesla V100} GPUs.

For t-SNE, we use the \textit{scikit-learn} \cite{scikit-learn} implementation with a perplexity of 10 and the remaining hyperparameters at their default values. We manually chose random seeds to produce the most visually clear images. We plot data using \textit{matplotlib} \citep{Hunter:2007}. We compute text similarity metrics using the \textit{nlg-eval} \citep{nlgeval} library, and compute tree edit distance using the \textit{zss} \footnote{\url{https://github.com/timtadh/zhang-shasha}} library.

We note that all software used in the development of this work is either in the public domain, open source, or does not specify a license.

\section{Additional Downstream Tasks}
\label{appendix-tasks}

\subsection{Headline Generation - OFEQ-10k}
\label{appendix-ofeq}

We examine the performance of MathGPT and GPT-2 on the headline generation task using the OFEQ-10k dataset \citep{mathsum}. We show the overall task results in Table \ref{tab:ofeq}, the expression-only task results in Table \ref{tab:ofeq-formula}, and the text-only task results in Table \ref{tab:ofeq-text}. We see surprisingly different results than on the EXEQ-300k dataset. We observe that MathGPT performs worse on the task overall than GPT-2 and GPT-2 Wiki, although it still outperforms them on the expression-only and text-only tasks. We also observe that, counterintuitively, GPT-2 Wiki performs slightly worse on the overall task than GPT-2, although it performs higher on the expression-only and text-only tasks. The negative impact of the Wikipedia pre-training, along with the fact that the trends are reversed when compared to the much larger EXEQ-300k dataset, lead us to believe that pre-training on a larger and more diverse dataset would improve performance on OFEQ-10k. We leave this investigation for future work.

\begin{table*}
    \centering
    \scalebox{.89}{
    \begin{tabular}{|l|c|c|c|}
        \hline
        \textbf{Model} & \textbf{BLEU-4} & \textbf{ROUGE-L} & \textbf{METEOR} \\
        \hline
        MathSum & $29.4$ & $39.0$ & $26.8$ \\
        \hline
        GPT-2 & $34.8 \pm 0.6$ & $46.9 \pm 0.5$ & $33.2 \pm 0.3$ \\
        GPT-2 Wiki & $34.5 \pm 0.6$ & $46.4 \pm 0.2$ & $32.8 \pm 0.3$ \\
        \hline
        MathGPT & $34.2 \pm 0.9$ & $47.2 \pm 0.4$ & $32.6 \pm 0.5$ \\
        \hline
    \end{tabular}}
    \caption{Results on headline generation for OFEQ-10k dataset.}
    \label{tab:ofeq}
\end{table*}

\begin{table*}
    \centering
    \scalebox{.89}{
    \begin{tabular}{|l|c|c|c|c|}
        \hline
        \textbf{Model} & \textbf{BLEU-4} & \textbf{ROUGE-L} & \textbf{METEOR} & \textbf{TED} \\
        \hline
        GPT-2 & $58.1 \pm 1.4$ & $74.0 \pm 0.3$ & $47.0 \pm 0.6$ & $3.806 \pm 0.066$ \\
        GPT-2 Wiki & $60.0 \pm 1.4$ & $76.2 \pm 0.3$ & $48.1 \pm 0.6$ & $3.444 \pm 0.092$ \\
        \hline
        MathGPT & *$62.2 \pm 0.8$ & *$76.8 \pm 0.1$ & *$49.1 \pm 0.3$ & *$2.862 \pm 0.031$ \\
        \hline
    \end{tabular}}
    \caption{Results on next expression prediction for OFEQ-10k dataset.}
    \label{tab:ofeq-formula}
\end{table*}

\begin{table*}
    \centering
    \scalebox{.89}{
    \begin{tabular}{|l|c|c|c|}
        \hline
        \textbf{Model} & \textbf{BLEU-4} & \textbf{ROUGE-L} & \textbf{METEOR} \\
        \hline
        GPT-2 & $15.1 \pm 0.5$ & $40.4 \pm 0.5$ & $20.2 \pm 0.5$ \\
        GPT-2 Wiki & $20.5 \pm 0.2$ & $38.3 \pm 0.4$ & $25.0 \pm 0.3$ \\
        \hline
        MathGPT & $20.6 \pm 0.5$ & *$43.7 \pm 0.2$ & *$26.6 \pm 0.3$ \\
        \hline
    \end{tabular}}
    \caption{Results on next text region prediction for OFEQ-10k dataset.}
    \label{tab:ofeq-text}
\end{table*}

\subsection{Student Answer Scoring}

We evaluate on the task of scoring student solutions to open-ended math problems from the ASSISTments system. This task helps assess the model's ability to apply mathematical reasoning to student data, as well as generalize to a classification setting. We use the same cleaned dataset and in-context meta-learning method as \citep{zhang2022automatic}. We also compare to the results from this work, which used a BERT model, and is the current state of the art on this dataset. Since this is a multi-label classification task we use a different set of metrics, specifically \textbf{Accuracy}, \textbf{F1}, macro-averaged area under the receiver operating characteristic curve (\textbf{AUC}), root mean squared error (\textbf{RMSE}) and Cohen's \textbf{Kappa}. We show the results of cross-validation on the task in Table \ref{tab:answer-scoring}. We observe that there is no significant difference between MathGPT and GPT-2 on this task. This is possibly due to the fact that many of the samples in the dataset either do not contain math expressions or contain only small ones, minimizing the effect of MathGPT's representations. The results may also imply that MathGPT sees most of its benefits in a generative rather than classification setting, although more experiments would need to be run to confirm this. We did not evaluate this task on GPT-2 Wiki. We note that the improvement over BERT is likely due to additional data processing we performed and small differences in our training setup.

\begin{table*}
    \centering
    \scalebox{.89}{
    \begin{tabular}{|l|c|c|c|c|c|}
        \hline
        \textbf{Model} & \textbf{Accuracy} & \textbf{F1} & \textbf{AUC} & \textbf{RMSE} & \textbf{Kappa} \\
        \hline
        BERT & -- & -- & $73.3 \pm 0.6$ & $1.077 \pm 0.002$ & $58.9 \pm 0.4$ \\
        \hline
        GPT-2 & $82.4 \pm 0.2$ & $61.8 \pm 0.5$ & $94.3 \pm 0.2$ & $0.933 \pm 0.007$ & $63.6 \pm 0.4$ \\
        \hline
        MathGPT & $82.2 \pm 0.3$ & $61.8 \pm 0.7$ & $94.2 \pm 0.1$ & $0.935 \pm 0.014$ & $63.7 \pm 0.5$ \\
        \hline
    \end{tabular}}
    \caption{Results on student answer scoring task.}
    \label{tab:answer-scoring}
\end{table*}

\section{Additional Qualitative Analysis}

\subsection{Additional Math Token Representations}
\label{appendix-tokens}

\begin{figure*}
\centering
  \fbox{\includegraphics[height=1.6in]{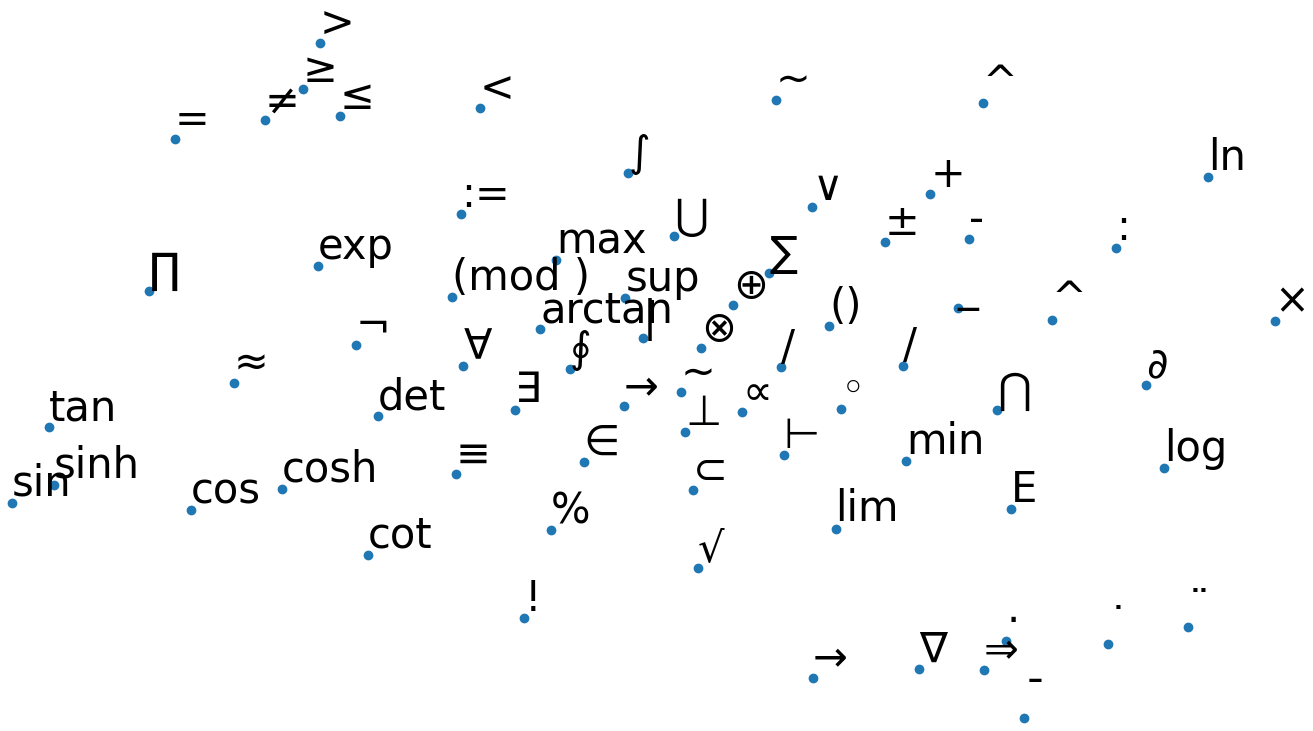}}
  \fbox{\includegraphics[height=1.6in]{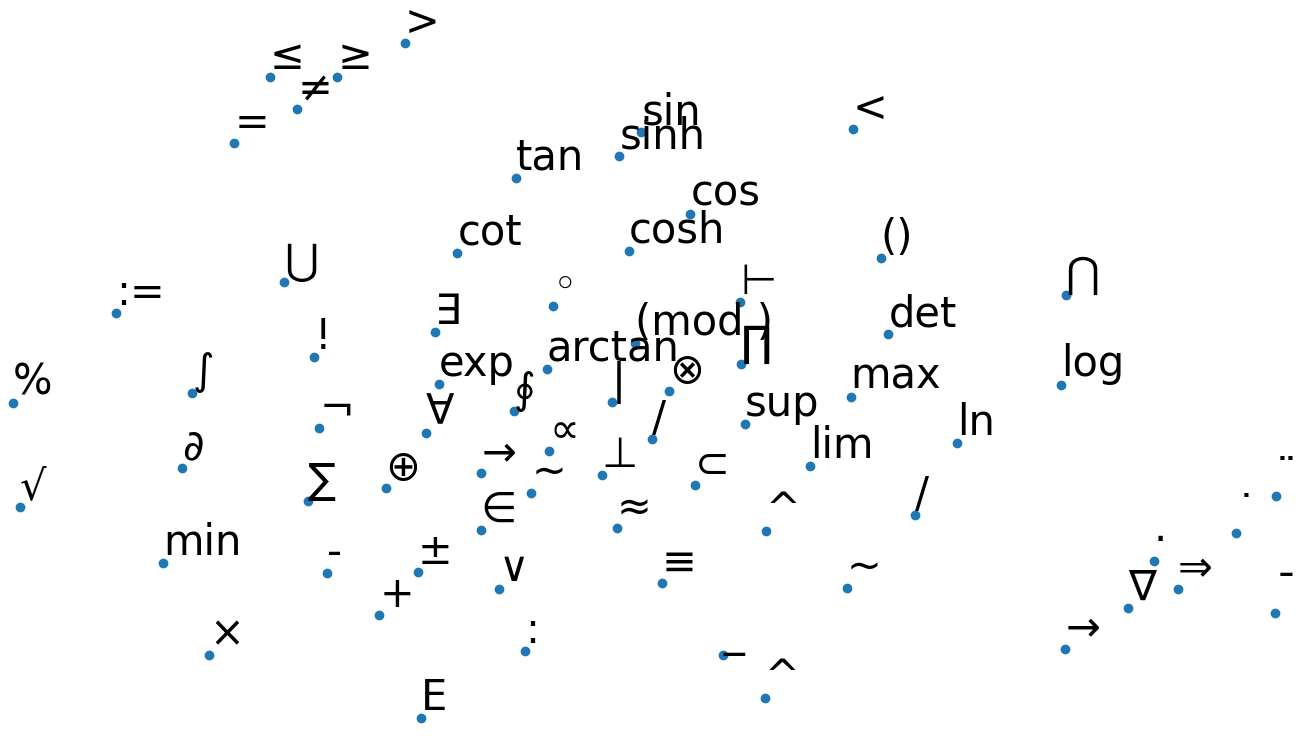}}
\caption{MathGPT operator token embeddings with different t-SNE random seeds.}
\label{fig:tokens-mathGPT-alt}
\end{figure*}

\begin{figure*}
\centering
  \fbox{\includegraphics[height=1.6in]{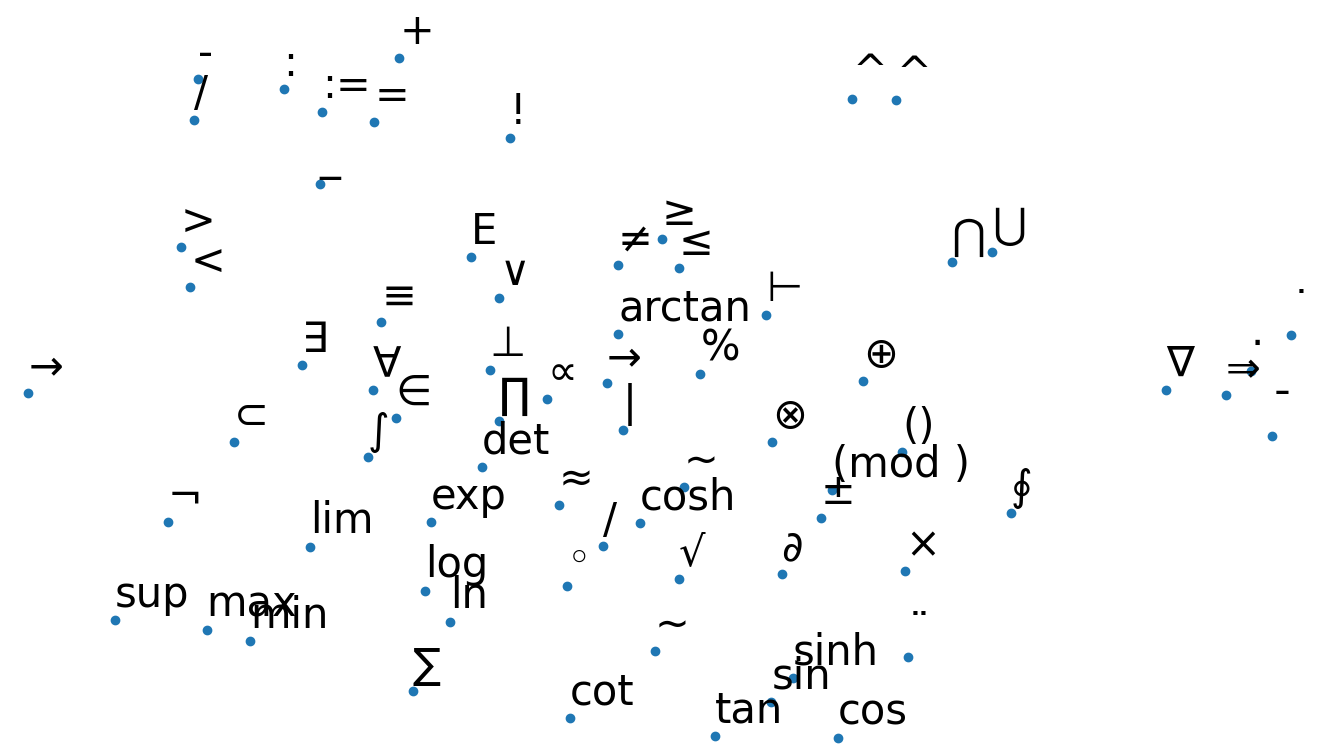}}
  \fbox{\includegraphics[height=1.6in]{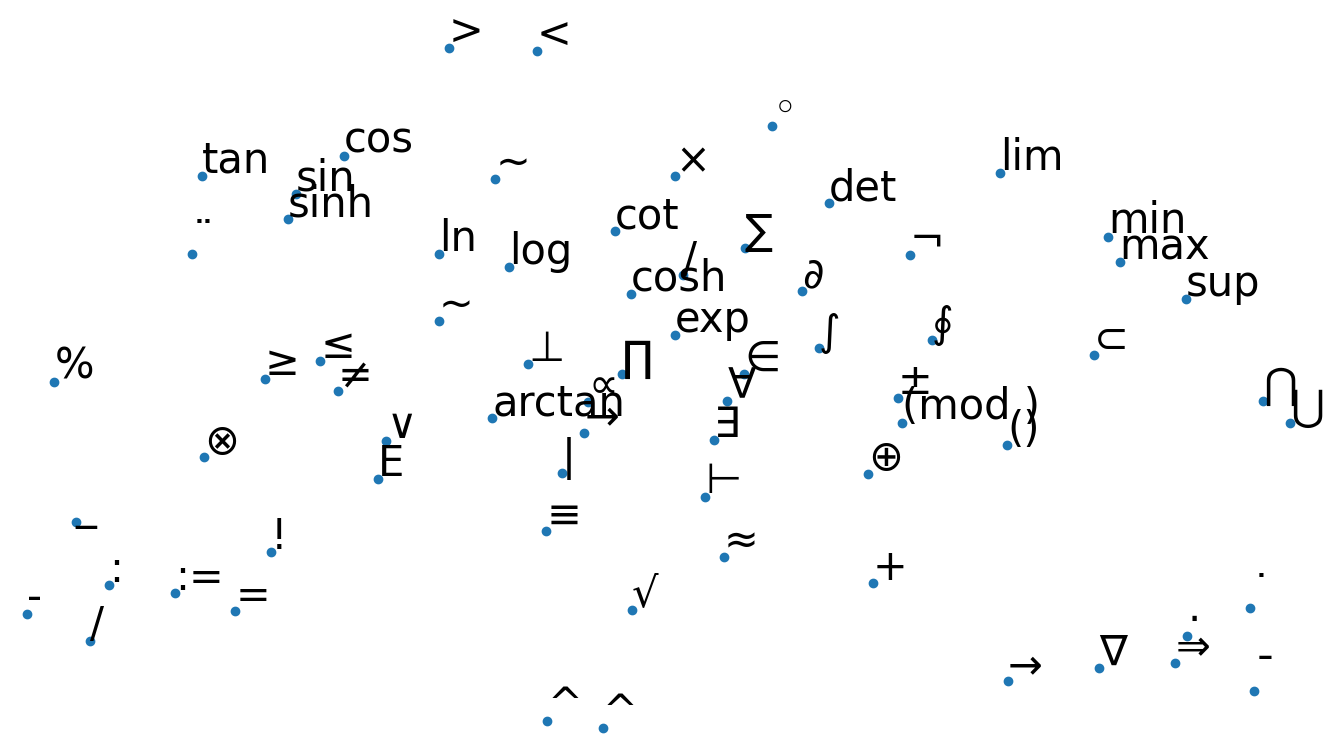}}
\caption{GPT-2 operator token embeddings with different t-SNE random seeds.}
\label{fig:tokens-gpt2-alt}
\end{figure*}

We examine the effects of different random seeds for t-SNE initialization on operator token representations. We show two such visualizations for MathGPT in Figure \ref{fig:tokens-mathGPT-alt} and two such visualizations for GPT-2 (fine-tuned on math Wikipedia articles) in Figure \ref{fig:tokens-gpt2-alt}. We observe that while most clusters stay the same across initializations, a few tokens tend to float around, in particular $\log$, $\ln$, and $\exp$.

\begin{figure*}
\centering
  \fbox{\includegraphics[height=1.6in]{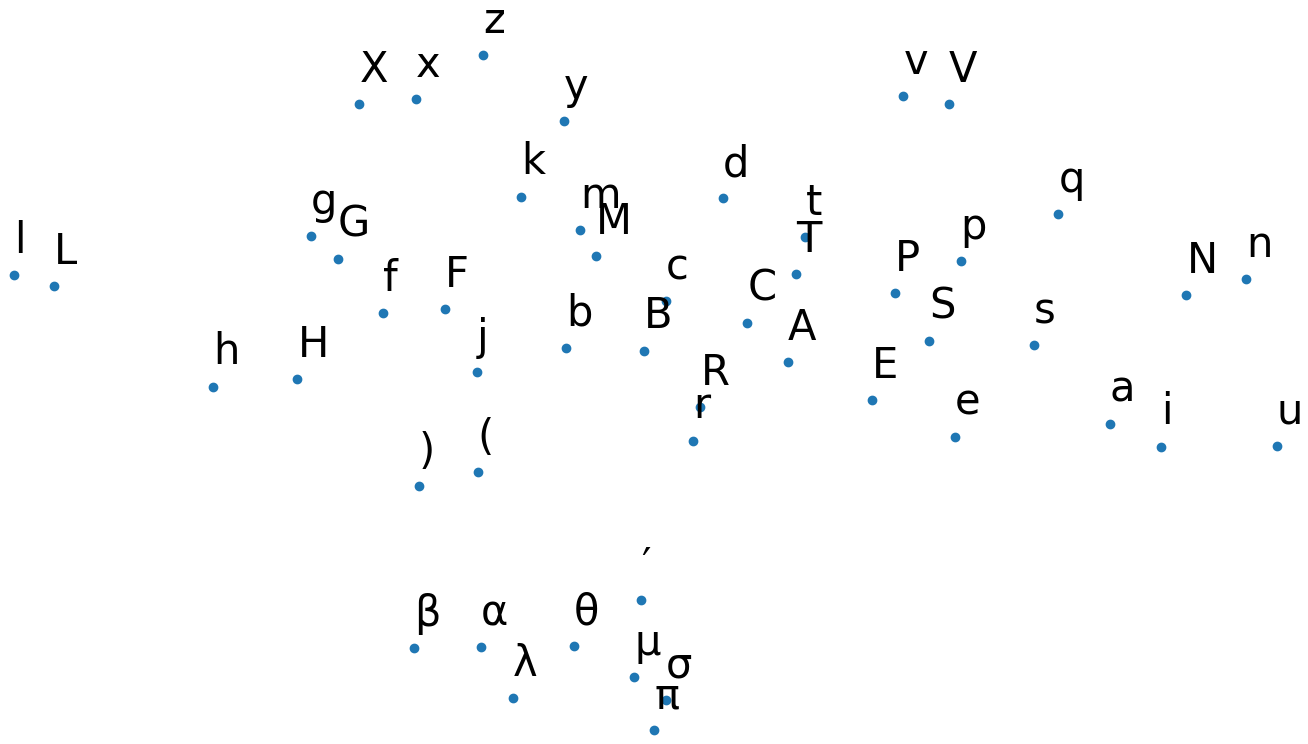}}
  \fbox{\includegraphics[height=1.6in]{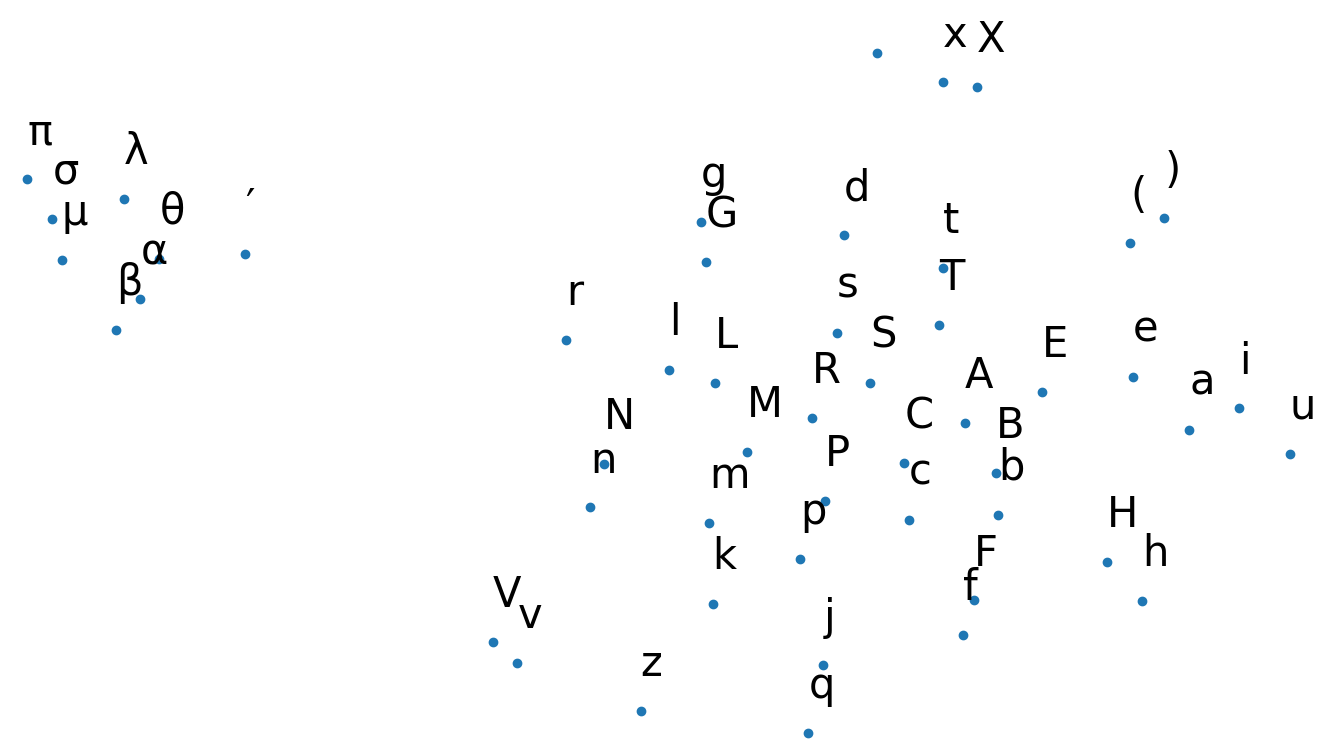}}
\caption{Variable token embeddings for MathGPT (left) and GPT-2 (right).}
\label{fig:tokens-vars}
\end{figure*}

We also show the representations of the 50 most common variable tokens, for both MathGPT and GPT-2, in Figure \ref{fig:tokens-vars}. For both models, we observe that lower- and upper-case versions of the same letter are close together, and that Greek letters are distant from the English letters. Interestingly, in contrast to operator tokens, there is very little change in variable token relationships across the models. This may be because the semantic meaning of variables is highly context-sensitive, preventing MathGPT from making generalizations at the token-level.

\section{Additional Dataset Details}
\label{appendix-datasets}

For completeness and transparency, we list the statistics and other details of all datasets used in this work. We list licenses when they are available, and privacy details when they are relevant.

The math Wikipedia articles used for pre-training are provided under a Creative Commons BY-SA license. We exclusively use the \textit{MathTagArticles} portion of the dataset, which contains 31,839 articles.

The EXEQ-300k and OFEQ-10k datasets consist of (train, validation, test) splits of sizes (261,341, 14,564, 14,574) and (10,301, 1,124, 1,123), respectively. Due to processing errors in a small portion of samples, the EXEQ-300k test set was reduced to a size of 14,474. However, we believe that this reduction is small enough ($\sim 0.7\%$) to not have a significant impact on reported results.

The Math23k dataset consists of 23,162 math word problems originally in Chinese and translated to English for use in this work. We chose to not use the publicly available test split for this dataset because it is very small compared to the size of the dataset (1000 samples), so cross-validation would provide a better measure of performance.

The Cognitive Tutor dataset consists of 8,298 unique problems and 95 students. All student identities are anonymized. Since student responses are constrained by the software, it is unlikely that they contain personally identifying information.

The version of the student answer scoring dataset we use consists of 1,333 unique problems and 130,940 responses, with each assigned a score from 1 to 5. The original dataset was introduced by \citep{erickson2020automated}. While student identities are anonymized, it is possible that personally identifying information is present in the open-ended student responses, and as such the dataset is not publicly available.

\section{Data Pipeline Illustration}
\label{pipeline-example}

\begin{figure*}[h]
    \centering
    \includegraphics[width=\linewidth]{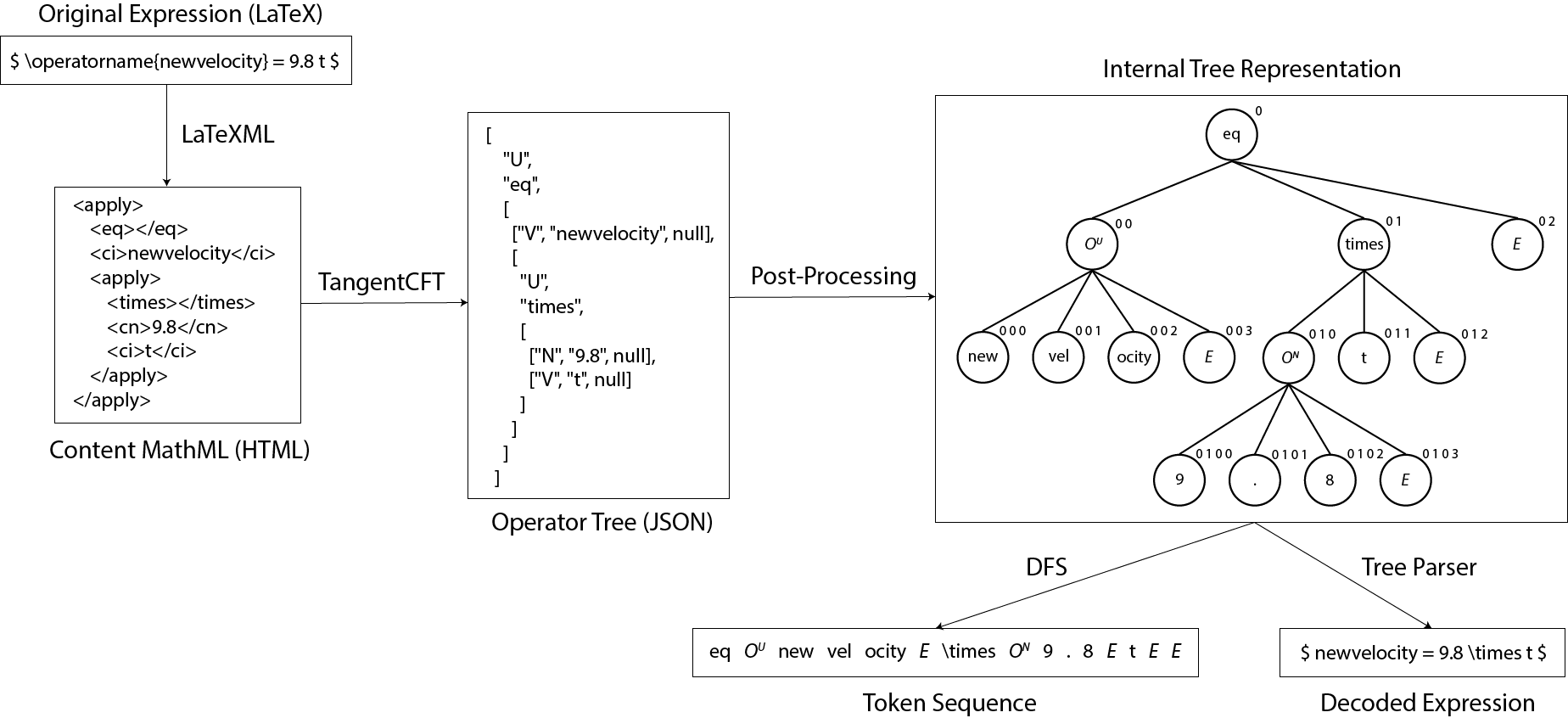}
    \caption{Data processing pipeline for the expression $\operatorname{newvelocity} = 9.8t$. The expression is initially converted to Content MathML format by LaTeXML, and is stored as HTML. It is then converted to a recursive operator tree format by TangentCFT, and is stored as JSON. Each node is represented by a 3-tuple, storing the TangentCFT type, followed by the node's name, followed by the list of children or \textit{null} if there are none. The expression is then sent through the post-processing pipeline, which tokenizes nodes, converts nodes out of the vocabulary to GPT-2-tokenized sub-trees, converts numbers to sub-trees, adds \textit{end} nodes, and computes tree position vectors (shown to the upper right of each node). This representation can be converted to a depth-first traversal of the tokens in order to be processed by MathGPT. It may also be converted back to human-readable text as LaTeX.}
    \label{fig:data_proc}
\end{figure*}

In Figure \ref{fig:data_proc}, we show the full data processing pipeline for a single expression.

\end{document}